\pdfoutput=1 % ensure pdflatex processing for arXiv

\def\year{2019}\relax
\documentclass[letterpaper]{article} %DO NOT CHANGE THIS
\usepackage{aaai19}  %Required
\usepackage{times}  %Required
\usepackage{helvet}  %Required
\usepackage{courier}  %Required
\usepackage{url}  %Required
\usepackage{graphicx}  %Required
\nocopyright

\newif\ifanonymous
\anonymousfalse

\ifanonymous
\else
\fi
\title{Constrained Exploration and Recovery from Experience Shaping}
\ifanonymous
    \author{Paper ID 6938 }
\else
    \author{Tu-Hoa Pham, Giovanni De Magistris, Don Joven Agravante, \\
    {\bf \Large Subhajit Chaudhury, Asim Munawar \and Ryuki Tachibana}\\ 
    IBM Research - Tokyo \\
    \texttt{\{pham,giovadem,subhajit,asim,ryuki\}@jp.ibm.com}, \texttt{don.joven.r.agravante@ibm.com}
    }
\fi

\usepackage[utf8]{inputenc} % allow utf-8 input
\usepackage[T1]{fontenc}    % use 8-bit T1 fonts
\usepackage{url}            % simple URL typesetting
\usepackage{nicefrac}       % compact symbols for 1/2, etc.
\usepackage{microtype}      % microtypography
\usepackage{soul}           % strikethrough
\usepackage{csquotes}
\usepackage{amsmath,amsthm,amsfonts,amssymb,amscd}
\usepackage{multirow,booktabs}
\usepackage[table]{xcolor}
\usepackage{fancyhdr}
\usepackage{mathrsfs}
\usepackage{wrapfig}
\usepackage{calc}
\usepackage{multicol}
\usepackage{cancel}
\usepackage{empheq}
\usepackage{framed}
\usepackage[inline]{enumitem}
\usepackage{siunitx}
\usepackage{algpseudocode}
\usepackage[caption=false, font=footnotesize]{subfig}
\theoremstyle{definition}

\newcommand{\Aparentheses}[1]{{\left( {#1} \right)}}
\newcommand{\Abraces}[1]{{\left\{ {#1} \right\}}}
\newcommand{\Abrackets}[1]{{\left[ {#1} \right]}}

\newcommand{\Anorm}[1]{{\left\lVert {#1} \right\rVert}}
\newcommand{\Areal}{\mathbb{R}}
\newcommand{\Ageneralineqvecel}{c}

\newcommand{\Aineqmatel}{G}
\newcommand{\Aineqmatlineel}{g}
\newcommand{\Aineqmatline}{\mathbf{\Aineqmatlineel}}
\newcommand{\Aineqmat}{\mathbf{\Aineqmatel}}
\newcommand{\Aineqvecel}{h}
\newcommand{\Aineqvec}{\mathbf{\Aineqvecel}}
\newcommand{\Aineqvecplus}{\Aineqvec_{+}}
\newcommand{\Aactionel}{x}
\newcommand{\Aaction}{\mathbf{\Aactionel}}
\newcommand{\Auncertainaction}{\widetilde{\Aaction}}
\newcommand{\Acorrectedaction}{\Aaction^{*}}
\newcommand{\Aactioni}[1]{\Aaction_{#1}}
\newcommand{\Astateel}{s}
\newcommand{\Astate}{\mathbf{\Astateel}}
\newcommand{\Astatei}[1]{\Astate_{#1}}
\newcommand{\Astatedomain}{S}
\newcommand{\Aactiondomain}{A}

\newcommand{\Areward}{r}
\newcommand{\Arewardi}[1]{\Areward_{#1}}
\newcommand{\Ahorizon}{T}
\newcommand{\Atransitionprobability}{P}
\newcommand{\Arewardfunction}{R}
\newcommand{\Ainitialstateprobability}{\rho_{0}}
\newcommand{\Adiscountfactor}{\gamma}
\newcommand{\Adiscountedexpectedreturn}{\eta}
\newcommand{\Apolicy}{\pi}

\newcommand{\Arecoverypolicy}{\Apolicy^{r}}
\newcommand{\Aexpectation}{\mathop{\mathbb{E}}}
\newcommand{\Astateactiontrajectory}{\tau}
\newcommand{\Aargmin}{\mathop{\text{argmin}}}

\newcommand{\Asa}{\Astate, \Aaction}

\newcommand{\Aps}{\Aparentheses{\Astate}}
\newcommand{\Apa}{\Aparentheses{\Aaction}}
\newcommand{\Apsa}{\Aparentheses{\Asa}}

\newcommand{\Anumobs}{{{n_\text{obs}}}}
\newcommand{\Anumact}{{{n_\text{act}}}}
\newcommand{\Anumineq}{{n_\text{in}}}

\newcommand{\Aineqindicesfull}{\left[ 1, \Anumineq \right]}

\newcommand{\Aloss}{\mathcal{L}}
\newcommand{\Ademonstration}{{d}}
\newcommand{\Ademonstrations}{\mathcal{D}}
\newcommand{\Anetwork}{\mathcal{N}}
\newcommand{\Apolicynetwork}{\Anetwork^{\Apolicy}}
\newcommand{\Adirectpolicynetwork}{\Anetwork^{\Apolicy}_{d}}
\newcommand{\Arecoverypolicynetwork}{\Anetwork^{\Apolicy}_{r}}
\newcommand{\Acnet}{\Anetwork^{C}}
\newcommand{\Auncertainexp}{^{*}}
\newcommand{\Agoodexp}{^{+}}
\newcommand{\Abadexp}{^{-}}

\newcommand{\Agooddemonstrations}{\Ademonstrations\Agoodexp}
\newcommand{\Abaddemonstrations}{\Ademonstrations\Abadexp}
\newcommand{\Auncertaindemonstrations}{\Ademonstrations\Auncertainexp}
\newcommand{\Adelta}{\delta}
\newcommand{\Adeltagood}{\Adelta\Agoodexp}
\newcommand{\Aindicator}{\Adeltagood}
\newcommand{\Aindicatorsa}{\Aindicator_{\Asa}}
\newcommand{\Adeltagoods}{\Adeltagood_{\Astate}}
\newcommand{\Adeltagoodsa}{\Adeltagoods\Apa}

\newcommand{\Asai}{\Astate, \Aaction, \Aindicatorsa}

\newcommand{\Apsai}{\Aparentheses{\Asai}}

\newcommand{\Amarginel}{M}

\newcommand{\Asatisfactionsymbol}{S}
\newcommand{\Aviolationsymbol}{V}

\newcommand{\Asatisfactionmarginel}{\Amarginel^{\Asatisfactionsymbol}}
\newcommand{\Aviolationmarginel}{\Amarginel^{\Aviolationsymbol}}

\newcommand{\Asatisfactionmarginvecisa}[1]{\Asatisfactionmarginel_{#1}{\Apsa}}
\newcommand{\Aviolationmarginvecisa}[1]{\Aviolationmarginel_{#1}{\Apsa}}

\newcommand{\Acnetloss}{\Aloss^{C}}
\newcommand{\Acnetlosssai}{\Acnetloss{\Apsai}}

\newcommand{\Aineqmats}{\Aineqmat^{\Astate}}

\newcommand{\Aineqvecs}{\Aineqvec^{\Astate}}
\newcommand{\Aineqvecpluss}{\Aineqvecplus^{\Astate}}
\newcommand{\Aineqveci}[1]{\Aineqvecel_{#1}}
\newcommand{\Aineqvecis}[1]{\Aineqveci{#1}^{\Astate}}

\newcommand{\Aineqmatlinei}[1]{\Aineqmatline_{#1}}
\newcommand{\Aineqmatlineis}[1]{\Aineqmatlinei{#1}^{\Astate}}
\newcommand{\Ageneralineqveci}[1]{\Ageneralineqvecel_{#1}}
\newcommand{\Ageneralineqvecis}[1]{\Ageneralineqvecel_{#1}^{\Astate}}
\newcommand{\Ageneralineqvecisa}[1]{\Ageneralineqvecis{#1}{\Apa}}
\DeclareMathOperator{\Arelu}{ReLU}

\newcommand{\Arewardrecovery}{\Areward_{r}}
\newcommand{\Arewardfail}{\Areward_{\text{fail}}}
\newcommand{\Arewardsuccess}{\Areward_{\text{goal}}}
\newcommand{\Arewarddistance}{\Areward_{\text{dist}}}
\newcommand{\Arewardalive}{\Areward_{\text{alive}}}
\newcommand{\Ainteriorpoint}{\hat{\Aaction}}

\newcommand{\Anumrecoverysteps}{n_{s}}
\newcommand{\Anumrecoveryattempts}{n_{a}}
\newcommand{\Aenvironment}{\mathcal{E}}
\newcommand{\Adirectenvironment}{\Aenvironment^{d}}
\newcommand{\Arecoveryenvironment}{\Aenvironment^{r}}
\newcommand{\Aexperiencereplaybuffer}{\mathcal{B}}
\newcommand{\Arltrajectory}{\tau_\text{PO}}
\newcommand{\Acnettrajectory}{\tau_{C}}

\begin{document}

\maketitle
\begin{abstract}
    We consider the problem of reinforcement learning under safety requirements, in which an agent is trained to complete a given task, typically formalized as the maximization of a reward signal over time, while concurrently avoiding undesirable actions or states, associated to lower rewards, or penalties. The construction and balancing of different reward components can be difficult in the presence of multiple objectives, yet is crucial for producing a satisfying policy. For example, in reaching a target while avoiding obstacles, low collision penalties can lead to reckless movements while high penalties can discourage exploration. To circumvent this limitation, we examine the effect of past actions in terms of safety to estimate which are acceptable or should be avoided in the future. We then actively reshape the action space of the agent during reinforcement learning, so that reward-driven exploration is constrained within safety limits. We propose an algorithm enabling the learning of such safety constraints in parallel with reinforcement learning and demonstrate its effectiveness in terms of both task completion and training time.
\end{abstract}

\section{Introduction}
\label{sec:introduction}

Recent work in reinforcement learning has established the potential
for deep neural network architectures to tackle difficult control and decision-making problems,
such as playing video games from raw pixel information~\cite{nature:mnih:2015},
Go~\cite{nature:silver:2016},
as well as robot manipulation~\cite{icra:haarnoja:2018}
and whole-body control~\cite{tog:peng:2018}.
Such problems are often characterized by
the high dimensionality of possible actions and observations,
making them difficult to solve
or even intractable for traditional optimization methods.
Still, deep reinforcement learning techniques have remained subject to
limitations including
poor sample efficiency,
large requirements in data and interactions with the environment
and strong dependency on an appropriately-designed reward signal~\cite{icml:duan:2016}.
In particular, a considerable challenge towards their applicability to real-world problems
is that of safety.
Indeed, while deep neural networks can reasonably be employed as \textit{black-box} controllers
within simulated or well-controlled environments,
limited interpretability and vulnerability to adversarial attacks~\cite{iclr:goodfellow:2015,arxiv:su:2017}
can hinder their deployment to situations where
poor decisions can have undesirable consequences for the agent or its environment,
e.g., in autonomous driving.
For such critical applications, 
deep neural networks can be used as a specialized service
for components offering stability and performance guarantees,
e.g., for visual recognition or dynamics prediction in conjunction with
model-predictive control~\cite{icra:williams:2017}.

\begin{figure}[!t]
    \centering
    \includegraphics[width=0.24\columnwidth]{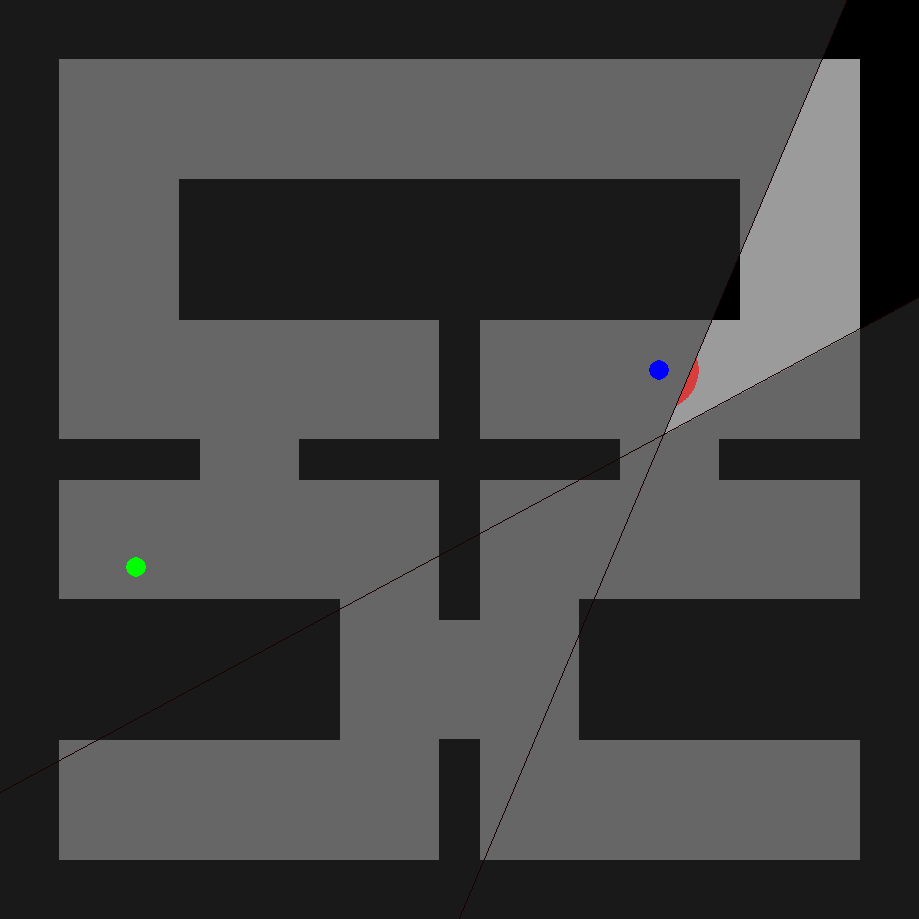}
    \includegraphics[width=0.24\columnwidth]{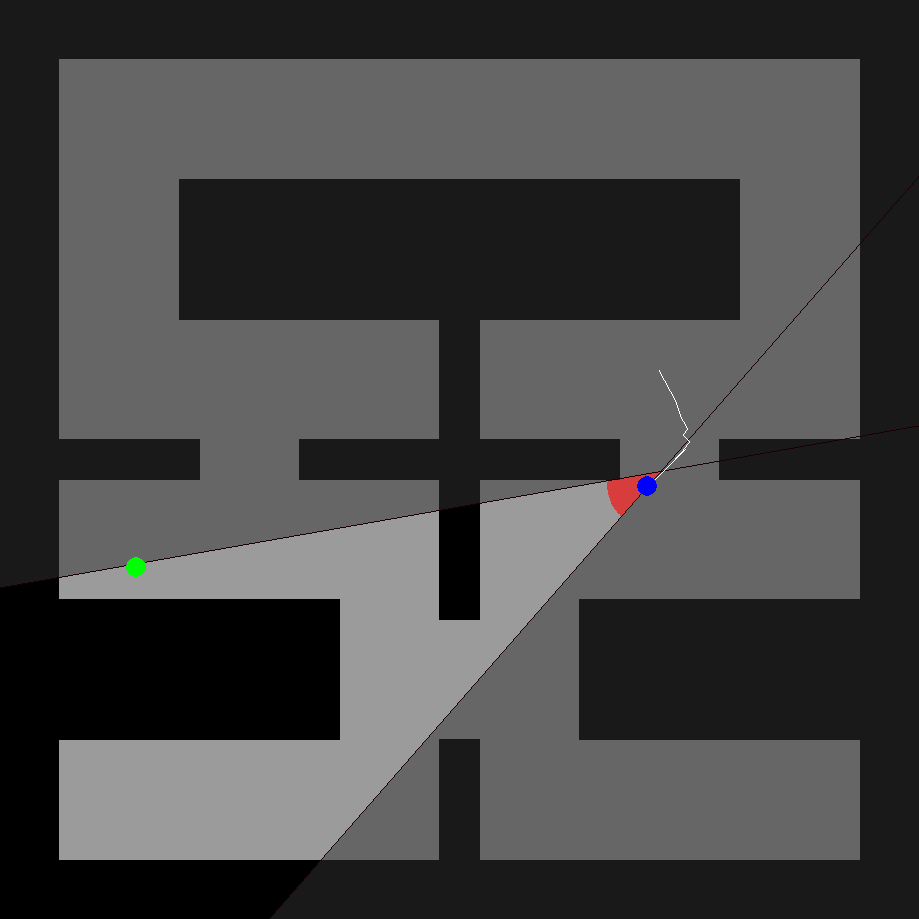}
    \includegraphics[width=0.24\columnwidth]{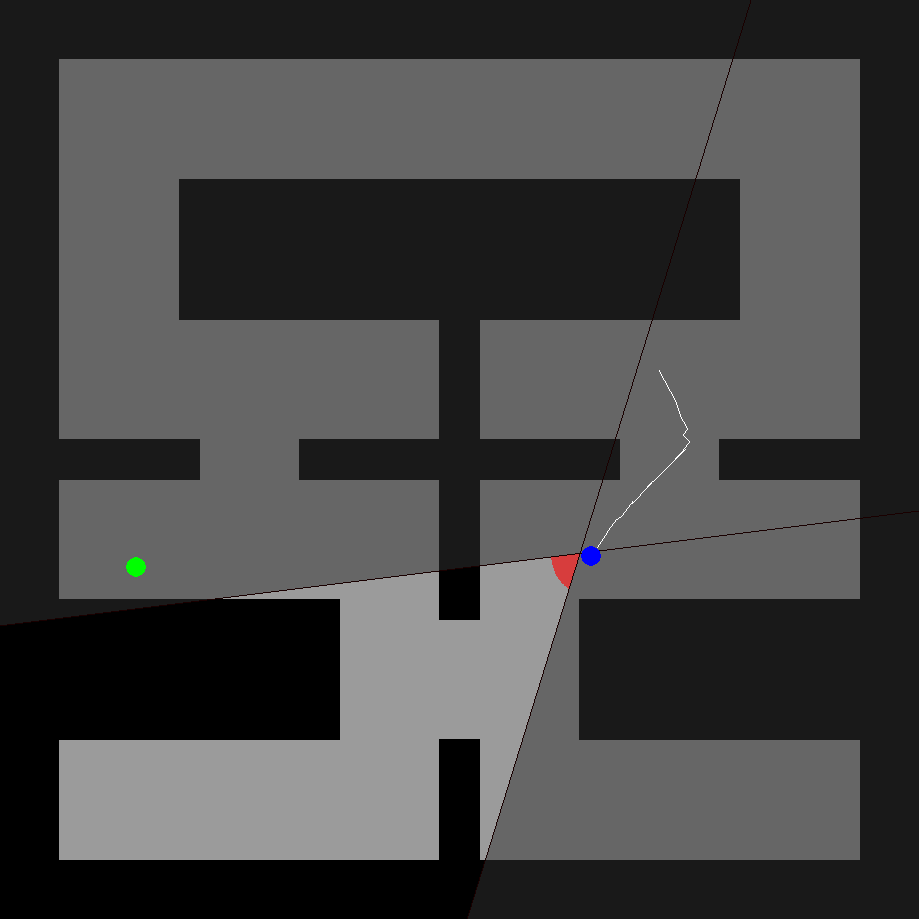}
    \includegraphics[width=0.24\columnwidth]{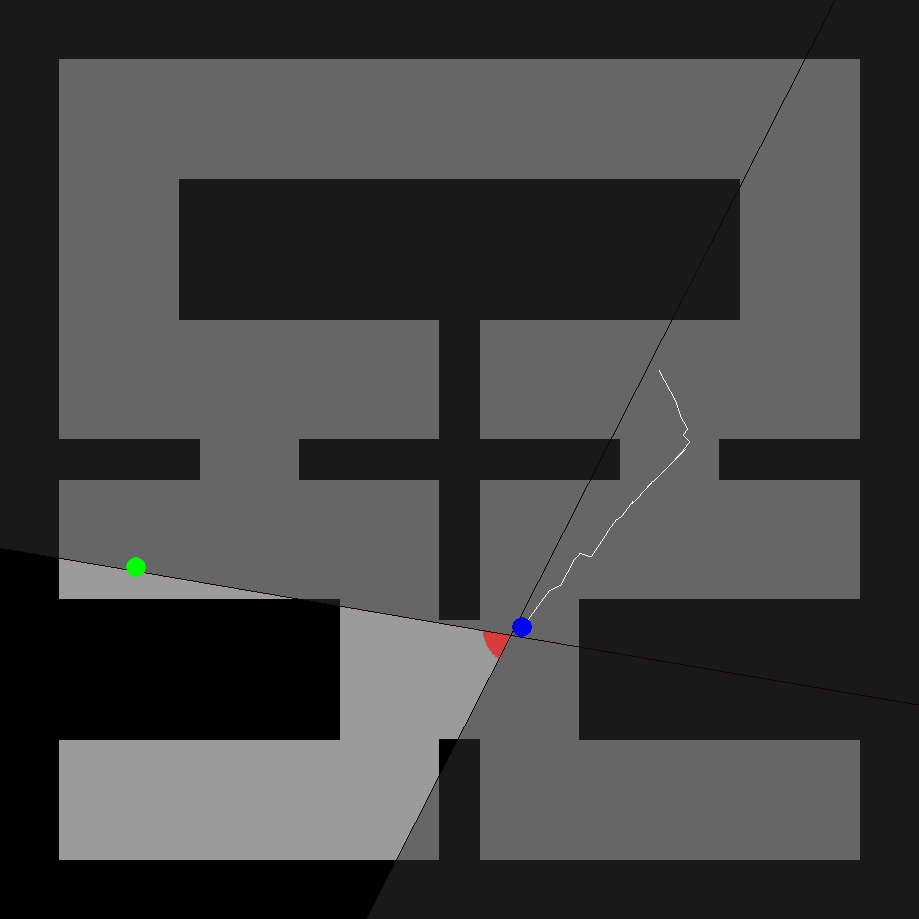} \\
    \includegraphics[width=0.24\columnwidth]{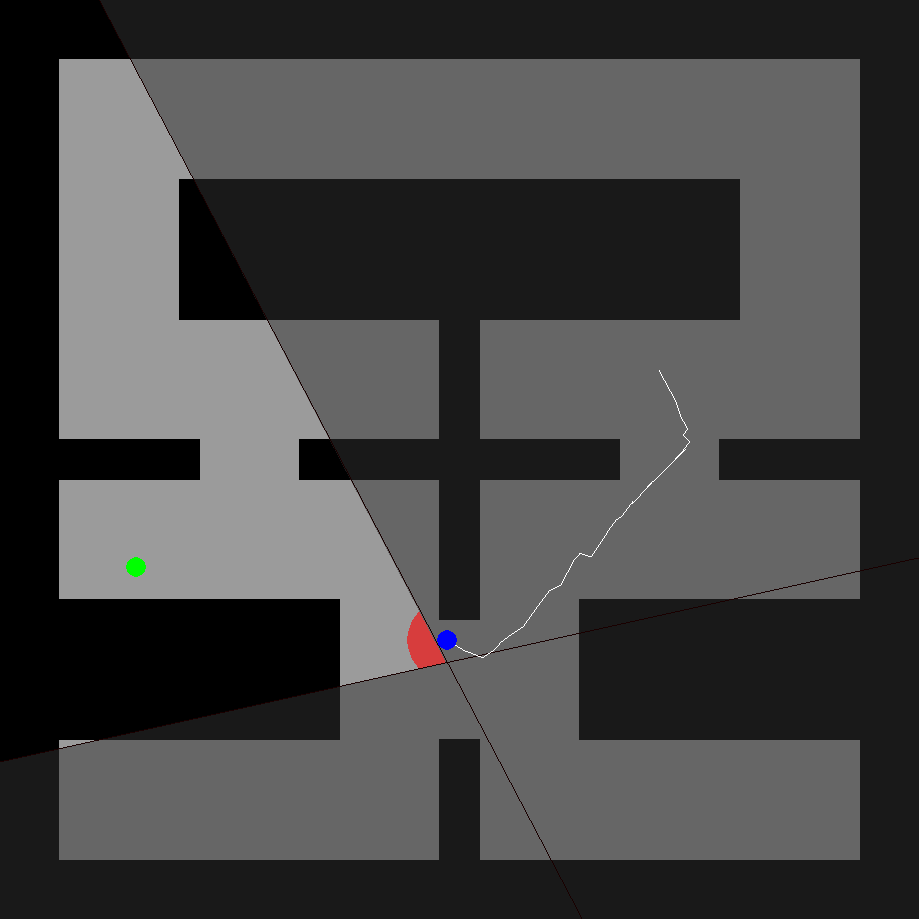}
    \includegraphics[width=0.24\columnwidth]{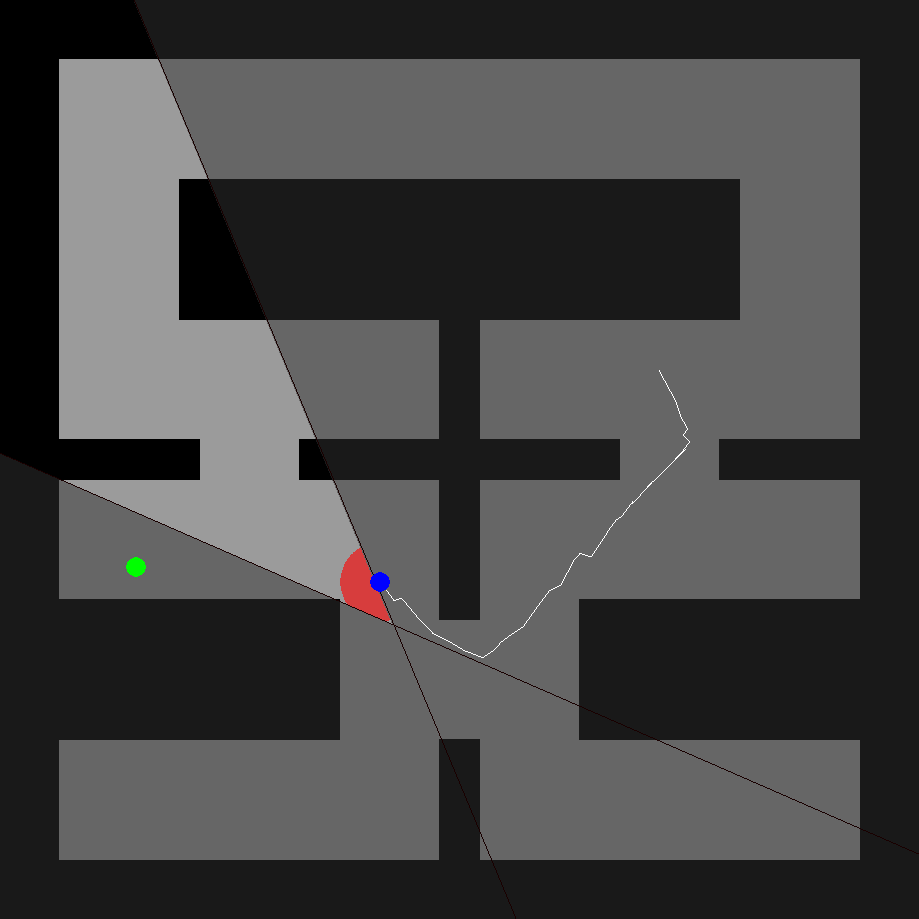}
    \includegraphics[width=0.24\columnwidth]{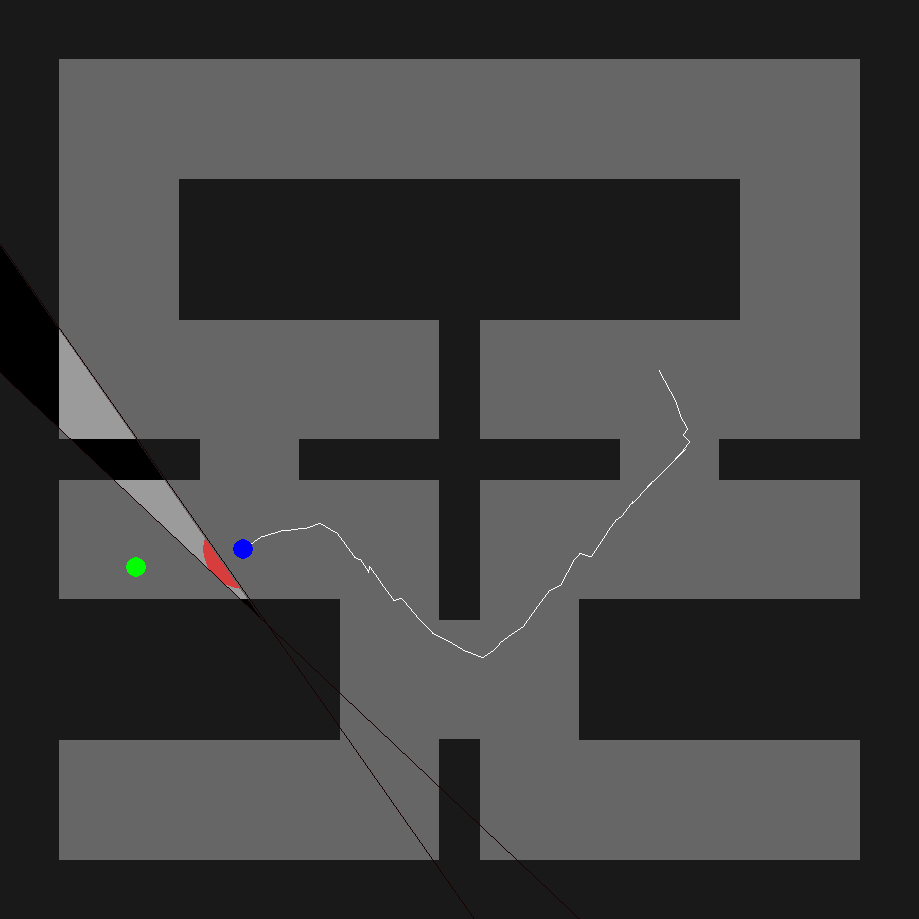}
    \includegraphics[width=0.24\columnwidth]{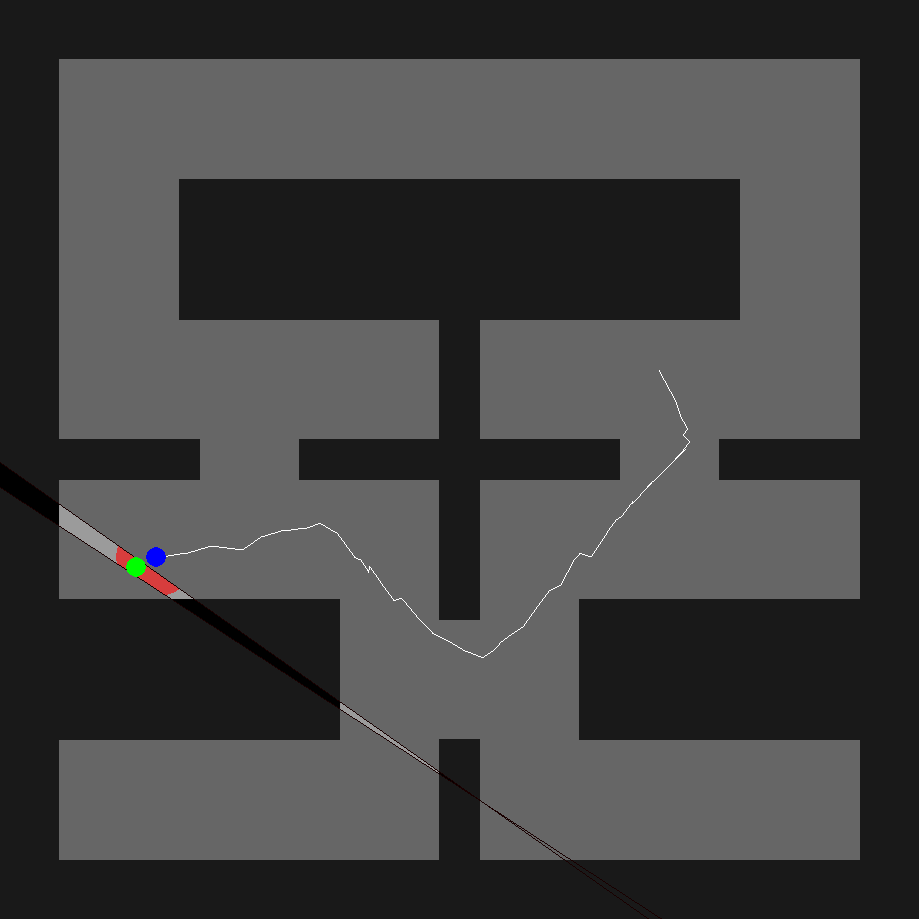}
    \caption{
        From left to right, top to bottom: consecutive actions taken
        during reinforcement learning in a maze environment with safety constraints.
        While in the unconstrained setting, the agent is permitted to take steps
        anywhere within the red circle, 
        our approach learns safety constraints (shaded area)
        preventing failure from hitting obstacles (black surfaces).
    }
    \label{fig:maze}
\end{figure}

In the absence of ground-truth physical models,
prediction robustness can also be improved when large amounts of data
are available or can be generated,
e.g.,
through transfer learning with domain randomization~\cite{iros:tobin:2017}.
Data can also be used to tackle the limitations of reinforcement learning
in terms of training time and reward crafting.
When reference trajectories are available,
e.g., demonstrated by an expert,
it is possible to initialize a control policy by behavioral cloning~\cite{nc:pomerleau:1991},
e.g., by training the neural network in a supervised manner
using reference states and actions as inputs-outputs.
However, behavioral cloning frequently requires tremendous amounts of data
that extensively span observation and action spaces.
Expert demonstrations were also used in interaction with the reinforcement
learning process to accelerate training~\cite{aaai:hester:2018}.
Alternatively, it is possible to use expert data
to infer a reward signal
that the expert is assumed to be following
through inverse reinforcement learning (IRL)~\cite{icml:ng:2000}.
However, IRL methods can perform poorly in the presence of imperfect demonstrations.
In addition, even when a reward signal can be estimated,
it can remain insufficient to train a control policy by reinforcement learning afterwards.
Towards this limitation,
\cite{nips:ho:2016} proposed to bypass the reward estimation step
by directly training a control policy together with a discriminator classifying
state-action pairs as expert-like or not,
in a manner analogous to generative adversarial networks~\cite{nips:goodfellow:2014},
showing successful imitation from very few expert trajectories
on robot control tasks.
Other successes have also been obtained
in meta-learning frameworks for generalization from single
demonstrations~\cite{nips:duan:2017}
or reinforcement learning from imperfect
demonstrations using multimodal policies~\cite{icml:haarnoja:2017,arxiv:gao:2018}.

In this work, we propose to use reference demonstrations in a novel manner:
not to train a control policy directly,
but rather to learn safety constraints towards the completion of a given task.
In this direction, we build upon the start of the art in safe reinforcement learning
(Section~\ref{sec:background}).
In contrast with traditional imitation learning frameworks,
our approach leverages both \textit{positive} and \textit{negative} demonstrations,
which we loosely define as aiming to complete and fail the designated task, respectively,
yet without need for optimality (e.g., maximum reward or fastest failure).
\begin{itemize}
    \item We demonstrate that
        it is possible to automatically learn action-space constraints in a supervised manner
        even when no ground-truth constraints are available,
        through the formulation of a loss function acting as a proxy
        for a convex optimization problem (Section~\ref{sec:learning_constraints}).
    \item Positive and negative reference demonstrations may not be available
        in many practical problems of interest.
        Thus, we derive an algorithm,
        Constrained Exploration and Recovery from Exploration Shaping (CERES),
        to discover both from
        parallel instantiations of a reinforcement learning problem
        while learning safety constraints
        (Section~\ref{sec:ceres}).
    \item On collision avoidance tasks with dynamics,
        we show that our approach makes reinforcement learning more efficient,
        achieving higher rewards in fewer iterations,
        while also enabling learning from reduced observations
        (Section~\ref{sec:experiments}).
\end{itemize}
Finally,
we discuss the challenges we encountered
and future directions for our work (Section~\ref{sec:conclusion}).
To facilitate its reproduction
and foster the research in constraint-based reinforcement learning,
we make our algorithms public and
open-source\footnote{\url{https://www.github.com/IBM/constrained-rl}}.

\section{Background and Motivation}
\label{sec:background}

\subsection{Reinforcement Learning}

We consider an infinite-horizon discounted Markov decision process (MDP)
characterized by:
$\Astatedomain$ a state domain representing observations
available to the agent we seek to control;
$\Aactiondomain$ an action domain representing how the agent can interact
with its environment;
$\Ainitialstateprobability: \Astatedomain \rightarrow [0,1]$
a probability distribution for the initial state;
$\Atransitionprobability: \Astatedomain \times \Aactiondomain \times \Astatedomain \rightarrow [0,1]$
a transition probability distribution describing how, from a
given state, taking an action can lead to another state;
and
$\Arewardfunction: \Astatedomain \times \Aactiondomain \times \Astatedomain \rightarrow \mathbb{R}$
a function associating rewards to such transitions.
With $\Adiscountfactor \in [0, 1)$ a discount factor on future reward expectations,
we aim to construct a stochastic policy
$\Apolicy: \Astatedomain \times \Aactiondomain \rightarrow [0, 1]$
that maximizes the $\Adiscountfactor$-discounted expected return
$\Adiscountedexpectedreturn\Aparentheses{\Apolicy}$:
\begin{align}
    \label{eq:discountedexpectedreturn}
    \Adiscountedexpectedreturn \Aparentheses{ \Apolicy } = 
    \Aexpectation\limits_{\tau}\Abrackets{
        \sum\limits_{i=0}^{\infty} {
            \Adiscountfactor^i \Arewardfunction\Aparentheses{
                \Astatei{i}, \Aactioni{i}, \Astatei{i+1}
            }
        }
    }%
    ,
\end{align}
with
$\Astateactiontrajectory = \Aparentheses{\Astatei{0}, \Aactioni{0}, \Astatei{1}, \Aactioni{1}, \dots}$
a sequence of states and actions
where the initial state $\Astatei{0}$ is initialized following $\Ainitialstateprobability$,
and each action $\Aactioni{i}$ is sampled following
$\Apolicy\Aparentheses{\cdot | \Astatei{i}}$
the control policy given the current state $\Astatei{i}$,
leading to a new state $\Astatei{i+1}$
following the transition function
$\Atransitionprobability\Aparentheses{\cdot | \Astatei{i}, \Aactioni{i}}$.
Through Eq.~\eqref{eq:discountedexpectedreturn},
we seek to maximize not a one-step reward,
but rather a reward expectation over time.
While $\Astatedomain, \Aactiondomain, \Ainitialstateprobability, \Atransitionprobability$
allow some variation in their implementation
(e.g., different resolutions for images as state space),
they remain
mostly characterized by the considered task.
In contrast, the reward function can often be engineered empirically,
from intuition, experience, and trial and error.
Such a process is ineffective and costly,
since evaluating a reward function candidate requires
training a policy with it.

We consider deep reinforcement learning in continuous action spaces,
in which actions are typically $\Anumact$-dimensional real-valued vectors,
$\Aaction \in \Aactiondomain \subset \Areal^\Anumact$
(e.g., joint commands for a robot arm).
Given an $\Anumobs$-dimensional input state vector
$\Astate \in \Astatedomain \subset \Areal^\Anumobs$,
actions are sampled following a neural network $\Apolicynetwork$
representing the control policy $\Apolicy$,
$\Aaction \sim \Apolicynetwork{\Aparentheses{\Astate}}$.
Multiple methods were developed to tackle such problems,
such as
Deep Deterministic Policy Gradient (DDPG)~\cite{arxiv:lillicrap:2015}
or Trust Region Policy Optimization (TRPO)~\cite{icml:schulman:2015},
which were benchmarked on robot control tasks in~\cite{icml:duan:2016}.
In this work, we build upon the Proximal Policy Optimization (PPO)
algorithm~\cite{arxiv:schulman:2017} and its 
OpenAI Baselines reference implementation~\cite{github:baselines:2017},
in which the control policy is an $\Anumact$-dimensional
multivariate Gaussian distribution of mean and standard deviation
predicted by the neural network $\Apolicynetwork$,
trained on-policy by interacting with the environment to collect
state-action-reward tuples
$\Aparentheses{\Astatei{i}, \Aactioni{i}, \Arewardi{i}}_{i=1,\Ahorizon}$
on a $\Ahorizon$-timestep horizon.
Alternative frameworks employing energy-based policies
also achieved significant results
on improved exploration and skill transfer
between tasks~\cite{icml:haarnoja:2017,icra:haarnoja:2018}.

\subsection{Safe Reinforcement Learning}
\label{sec:safe_rl}

While failure is most often permissible in simulated environments,
real-world applications often come with requirements in terms of safety,
for both the artificial agent and its environment.
Indeed, poor decisions may have undesirable consequences,
both in the physical world
(e.g., an autonomous vehicle colliding with another vehicle or person)
and within information systems (e.g., algo trading).
Thus, the topic of safe reinforcement learning has been the subject of considerable research
from multiple perspectives~\cite{jmlr:garcia:2015}.
From the deep reinforcement learning domain,
\cite{icml:achiam:2017} recently proposed a trust region method,
named Constrained Policy Optimization (CPO),
enabling the training of control policies
with near-satisfaction of given, known safety constraints.
Towards real-world applications,
\cite{icra:pham:2018} proposed to combine the TRPO reinforcement learning algorithm
with an optimization layer that takes as input an action predicted by a neural network
policy and correct it to lie within safety constraints
via convex optimization.
There again, safety constraints are required to be specified in advance.
Namely, given a state $\Astate$, an action is sampled from a neural network policy
as $\Auncertainaction \sim \Apolicynetwork{\Aps}$.
Instead of directly executing $\Auncertainaction$ onto the environment,
as the neural network has no explicit safety guarantee,
it is first corrected by solving the following
quadratic program (QP)~\cite{oe:mattingley:2012}:
\begin{align}
    \label{eq:qp}
    \Acorrectedaction =
    \Aargmin\limits_{\Aaction \in \Areal^{\Anumact}}
    \Abraces{
        \Anorm{\Aaction - \Auncertainaction}^2
        \enskip \text{such that} \enskip
        \Aineqmat\Aaction \leq \Aineqvec
    }
    ,
\end{align}
with $\Aineqmat$ and $\Aineqvec$ linear constraint matrices
of respective size
$\Anumineq \times \Anumact$ and $\Anumineq \times 1$
describing the range of possible actions.
The closest action satisfying these constraints, $\Acorrectedaction$,
is then executed in the environment.
While in Eq.~\eqref{eq:qp},
$\Aineqmat$ and $\Aineqvec$
are assumed provided by the user,
in our work, we instead propose to learn them.
This is a rather unexplored idea,
since safety constraints
can sometimes be constructed in a principled way,
e.g., using the equations of physics in robotics.
However, doing so is often cumbersome and possibly imprecise,
as it depends on the availability of an accurate model of the agent
and its environment.
In contrast, our approach operates in a complete model-free fashion
and is able to learn from direct demonstrations
(possibly generated from scratch),
without need for prior knowledge.

Other evidence of reinforcement learning acceleration through improved exploration
was presented in~\cite{aaai:wachi:2018},
where safety constraints where modelled with Gaussian processes.
From the perspective of planning,
\cite{aaai:cserna:2018} defined safety not as a numerical quantity as in the previous works,
but through the notion of avoiding dead ends,
from which the task can no longer be completed.
In our work,
we similarly define \emph{negative} demonstrations $\Ademonstration \in \Abaddemonstrations$
as state-action
couples $\Ademonstration = \Aparentheses{\Asa}$ such that taking $\Aaction$
from $\Astate$ inevitably leads to failure:
either directly
(e.g., the agent immediately crashes against a wall)
or because recovery is no longer possible after taking $\Aaction$
(e.g., still accelerating despite passing a minimum braking distance).
Conversely, we define \emph{positive}
demonstrations $\Ademonstration \in \Agooddemonstrations$
as state-action couples
such that the agent can still recover from the resulting state
(e.g., starting to decelerate before passing the minimum braking distance).
The set of demonstrations $\Auncertaindemonstrations$
that are neither known to be positive or negative are called \emph{uncertain}.
Although
determining beyond doubt whether an uncertain demonstration is positive or negative
may be intractable in many cases,
we propose a heuristic approach to sample and classify such demonstrations
through a specialized reinforcement learning process.
Our approach is thus related to that of~\cite{iclr:eysenbach:2018},
where a reset policy was learned to return the environment to a safe state
for future attempts.
In~\cite{icml:pinto:2017},
increased robustness was achieved by training the control policy together
with an adversary learning to produce optimal perturbations.
While such perturbations can be used as negative demonstrations,
we seek to collect a variety of such examples without need for optimality
(e.g., any action leading an agent to collide against a wall,
without necessary inducing the greatest impact).
Finally,
although not reinforcement learning,
we were also inspired by the work of
\cite{iros:gandhi:2017},
where negative demonstrations were collected
by purposely crashing a drone into surrounding
objects to learn whether a direction is safe to fly to
as a simple binary classification problem.

\section{Learning Action-Space Constraints from Positive and Negative Demonstrations}
\label{sec:learning_constraints}

\subsection{Definitions}

\paragraph{State-dependent action-space constraints}
Let $\Aparentheses{\Ageneralineqveci{i}}_{i\in\Aineqindicesfull}$
denote a set of $\Anumineq$ constraints functions operating on
actions $\Aaction \in \Areal^\Anumact$,
of the general form:
\begin{align}
    \label{eq:general_individual_constraints}
    \Ageneralineqveci{i}{\Apa} \leq 0, \quad i \in \Aineqindicesfull
    .
\end{align}
In general, the constraint functions $\Ageneralineqveci{i}$ can take different forms
but are typically real-valued,
e.g., $\Ageneralineqveci{i}{\Apa} = \Anorm{\Aaction}_{2} - 1$,
the second-order cone inequality constraining $\Aaction$ to be of
$\mathcal{L}^2$ norm 1 or less.
We consider in particular the case of linear inequalities,
e.g., $2 \Aactionel_0 - \Aactionel_1 \leq 3$,
parameterized by 
a row vector $\Aineqmatlinei{i}$ of size $\Anumact$
and a scalar $\Aineqveci{i}$
such that:
\begin{align}
    \label{eq:linear_constraints}
    \Ageneralineqveci{i}{\Apa} = \Aineqmatlinei{i} \Aaction - \Aineqveci{i},
    \quad i \in \Aineqindicesfull
    .
\end{align}
With
$\Aineqmat = \left[ \Aineqmatlinei{1}, \dots, \Aineqmatlinei{\Anumineq} \right]^T$
and
$\Aineqvec = \left[ \Aineqveci{1}, \dots, \Aineqveci{\Anumineq} \right]^T$
the constraint matrices of respective size
$\Anumineq \times \Anumact$ and $\Anumineq \times 1$,
Eq.~\eqref{eq:general_individual_constraints}
takes the familiar form
$\Aineqmat \Aaction \leq \Aineqvec$
of Eq.~\eqref{eq:qp},
with inequalities considered row-wise.
We are interested in estimating such constraint matrices
as functions of state vectors $\Astate \in \Areal^\Anumobs$,
e.g., as outputs of a neural network $\Acnet$:
\begin{align}
    \label{eq:constraints_from_state}
    \Aineqmats \Aaction \leq \Aineqvecs, \enskip
    \text{with} \enskip \Aparentheses{\Aineqmats, \Aineqvecs} = \Acnet{\Aps}
    .
\end{align}
Formally,
we thus consider constraints that operate on the action domain and depend on the current state
(e.g., an autonomous vehicle may not accelerate more than a given rate
-- action constraint -- if another vehicle is less than a given distance ahead -- current state).
Eq.~\eqref{eq:linear_constraints}
can thus be rewritten:
\begin{align}
    \label{eq:linear_constraints_state}
    \Ageneralineqvecis{i}{\Apa} = \Aineqmatlineis{i} \Aaction - \Aineqvecis{i},
    \quad i \in \Aineqindicesfull
    ,
\end{align}
with $\Ageneralineqvecis{i}{\Apa} \leq 0$ when the constraint is satisfied,
and $\Ageneralineqvecis{i}{\Apa} > 0$ when it is violated.
Given a demonstration $\Ademonstration = \Apsa$,
wethen  define, for each constraint $i \in \Aineqindicesfull$,
a satisfaction margin $\Asatisfactionmarginvecisa{i}$
and a violation margin $\Aviolationmarginvecisa{i}$:
\begin{align}
    \label{eq:satisfaction_margin}
    \Asatisfactionmarginvecisa{i}
    = \max\Aparentheses{0, -\Ageneralineqvecisa{i}}
    = \Arelu\Aparentheses{-\Ageneralineqvecisa{i}}, \\
    \label{eq:violation_margin}
    \Aviolationmarginvecisa{i}
    = \max\Aparentheses{0, \Ageneralineqvecisa{i}}
    = \Arelu\Aparentheses{\Ageneralineqvecisa{i}},
\end{align}
with $\max$ the maximum operator, which for comparisons with zero
can be represented by $\Arelu$ the rectified linear unit.
Thus,
 $\Asatisfactionmarginvecisa{i}$
(resp. $\Aviolationmarginvecisa{i}$),
is positive if
the $i$-th constraint 
is satisfied (resp. violated),
and zero otherwise.
Finally,
given a set of known positive and bad demonstrations,
we associate to each an indicator
$\Aindicatorsa$ equal to $1$ if $\Apsa$ is a positive demonstration
and $0$ otherwise.

\subsection{Constraint Training Loss}

In this Section,
we assume the availability of
state-action demonstrations $\Apsa$ along with associated indicators $\Aindicatorsa$
(e.g., provided by a human expert).
We then seek to construct constraint functions
$\Aparentheses{\Ageneralineqveci{i}}_{i\in\Aineqindicesfull}$
that satisfy the following.
If $\Apsa$ is a positive demonstration,
then we want \emph{all} constraints to be satisfied:
\begin{align}
    \label{eq:good_def_base}
    \Aindicatorsa = 1
    \implies
    \forall i \in \Aineqindicesfull,
    \Ageneralineqvecisa{i} \leq 0
    .
\end{align}
Having all constraints satisfied is equivalent to having none violated.
Using the violation margin defined in Eq.~\eqref{eq:violation_margin}
yields:
\begin{align}
    \label{eq:good_def_intermediate}
    \Aindicatorsa = 1
    \implies
    \forall i \in \Aineqindicesfull,
    \Aviolationmarginvecisa{i} = 0
    .
\end{align}
Since by definition, all margins are non-negative, we get:
\begin{align}
    \label{eq:good_def_final}
    \Aindicatorsa = 1
    \implies
    \max\limits_{i \in \Aineqindicesfull}
    \Abraces{
    \Aviolationmarginvecisa{i}
    }
    = 0
    .
\end{align}

Conversely,
if $\Apsa$ is a negative demonstration,
we want \emph{at least one} constraint to be violated:
\begin{align}
    \label{eq:bad_def_base}
    \Aindicatorsa = 0
    \implies
    \exists i \in \Aineqindicesfull,
    \Ageneralineqvecisa{i} > 0
    .
\end{align}
This amounts to having at least one constraint of zero satisfaction margin,
while others can be strictly positive:
\begin{align}
    \label{eq:bad_def_final}
    \Aindicatorsa = 0
    \implies
    \min\limits_{i \in \Aineqindicesfull}
    \Abraces{
        \Asatisfactionmarginvecisa{i}
    }
    = 0
    .
\end{align}

Thus, we can define a constraint loss $\Acnetloss$
comprising the maximum violation for positive demonstrations following
Eq.~\eqref{eq:good_def_final}
and the mimimum satisfaction for negative demonstrations following
Eq~\eqref{eq:bad_def_final}:
\begin{align}
    \label{eq:full_loss}
    \begin{aligned}
        \Acnetlosssai & = 
        \Aindicatorsa
        \max\limits_{i \in \Aineqindicesfull}
        \Abraces{\Aviolationmarginvecisa{i}} \\
        & +
        \Aparentheses{
            1 - \Adeltagoodsa
        }
        \min\limits_{i \in \Aineqindicesfull}
        \Abraces{\Asatisfactionmarginvecisa{i}}
    \end{aligned}
\end{align}
Backtracing from Eq.~\eqref{eq:full_loss} to Eq.~\eqref{eq:constraints_from_state}
shows that $\Acnetlosssai$ is computed as a succession of differentiable operations
from $\Apsai$.
As the constraint matrices are computed in particular from $\Astate$
being fed through the constraint network $\Acnet$,
it can thus be trained in a supervised manner by
minimizing $\Acnetloss$ as a training loss,
using existing stochastic optimization methods such as 
Adam~\cite{arxiv:kingma:2014}.
Still, some considerations remain.

\subsection{Optimizing the Constraint Loss in Practice}

\subsubsection{Constraint Normalization}

In practice, directly minimizing the loss function of Eq.~\eqref{eq:full_loss}
does not suffice to yield useful constraints in practice.
Indeed, from the definition of satisfaction and violation margins,
it appears that $\Aineqmats = \mathbf{0}$
and $\Aineqvecs = \mathbf{0}$
yields a trivial minimum for $\Acnetloss$.
In fact, simply having $\Aineqmats = \mathbf{0}$ results in
the optimization problem being ill-defined.
Considering individual constraint parameters
$\Aparentheses{\Aineqmatlineis{i}, \Aineqvecis{i}}$,
we can instead observe that when $\Aineqmatlineis{i}$ is non-zero,
$\Aineqmatlineis{i}\Aaction - \Aineqvecis{i} = 0$ is the equation of a hyperplane in
$\Areal^{\Anumact}$ (i.e., a line in 2D action space, a plane in 3D action spaces, etc.),
of normal $\Aineqmatlineis{i}$ itself.
Geometric considerations then yield that
$\frac{\Aineqmatlineis{i}\Aaction - \Aineqvecis{i}}{\Anorm{\Aineqmatlineis{i}}}$
is the signed distance between $\Aaction$ and the constraint hyperplane.
It thus appears that having each row $\Aineqmatlineis{i}$
of the predicted constraint matrix $\Aineqmats$ to be of unit norm
would be practical,
for both avoiding trivial optima while maintaining geometric interpretability.
One possibility is to systematically renormalize
satisfaction and violation margins by division with 
the norm of each $\Aineqmatlineis{i}$ post-prediction,
within Eqs.~\eqref{eq:satisfaction_margin} and~\eqref{eq:violation_margin}.
However, we noted that doing so could result in two problems in particular:
the neural network predictions growing indefinitely large as they are normalized within
the training loss,
or conversely decreasing in norm such that $\Anorm{\Aineqmatlineis{i}}$
eventually becomes close to zero, causing numerical errors.

\paragraph{Unit constraint matrices}
Instead of re-normalizing row constraint matrices \emph{a posteriori},
we adopt an alternative formulation ensuring that they are of unit norm in the first place.
Recalling that each row can be interpreted as a unit vector in $\Areal^\Anumact$,
we have the constraint neural network predict it
in generalized spherical coordinates,
representing $\Anumact$-dimensional vectors in Cartesian coordinates
as a radius $r$ and $\Anumact - 1$ angular coordinates
$\phi_1, \dots, \phi_{\Anumact - 1}$.
For example, 2D vectors in Cartesian coordinates can be computed from polar coordinates
$\Aparentheses{r, \phi}$ as
$x_0 = r \cos{\Aparentheses{\phi}}, x_1 = r \sin{\Aparentheses{\phi}}$,
with analogous formulas for generalized $\Anumact$-dimensional spheres.
By simply setting the radius to $1$,
any combination of angles in $\Areal^{\Anumact-1}$
produces in a unit vector in $\Areal^\Anumact$.
We then change the output layer of the neural network $\Acnet$ so that it predicts
$\Anumact$ parameters for each constraint $i$:
$\Anumact-1$ spherical coordinates for $\Aineqmatlinei{i}$
and the scalar $\Aineqveci{i}$.
The transformation from spherical to Cartesian coordinates only involving
cosine and sine functions, the differentiability of the loss function is preserved.

\paragraph{Avoiding constraint incompatibility}

While constraint satisfaction and violation terms appear together in Eq.~\eqref{eq:full_loss},
they may not be optimized on partially overlapping demonstrations,
e.g., a positive and negative demonstration sharing the same state
(one action leading to failure and the other not).
As isolated demonstrations do not suffice to cleanly separate action-spaces
for any state,
it is possible that the neural network
produces constraints that
minimize the training loss $\Acnetloss$
but
are incompatible with each other.
For example, it is not possible to simultaneously satisfy
$x_0 <= 1$ and $x_0 >= 2$.
Instead, we would like to ensure that the domain described by
$\Aineqmats \Aaction \leq \Aineqvecs$ never boils down to the empty set.
Remark that if $\Aineqvecs \geq \mathbf{0}$,
then the optimization problem is always solvable,
since the valid domain now contains at least $\Aaction = \mathbf{0}$.
While it is straightforward to enforce $\Aineqvecs \geq \mathbf{0}$,
e.g., by passing it through a $\Arelu$ operation,
having $\Aaction = \mathbf{0}$ as default fallback action
may not always be safe in practice.
Instead of $\mathbf{0}$,
given an arbitrary point $\Ainteriorpoint \in \Areal^{\Anumact}$,
it is possible to parameterize constraints such that $\Ainteriorpoint$
always satisfies
$\Aineqmats \Aaction \leq \Aineqvecs$,
by decomposing the right hand-side into
$\Aineqvec = \Aineqmats \Ainteriorpoint + \Aineqvecpluss$,
with $\Aineqvecpluss \geq \mathbf{0}$.
While $\Ainteriorpoint$ can be fixed manually,
it can also be considered as an \emph{interior point} that can be learned
and shared with each individual constraint.
Finally, the bounds of $\Aineqvecpluss$ can also be set
to guarantee,
e.g., a minimum or maximum distance between constraints
and the interior point $\Ainteriorpoint$.
In the following, we set the minimum value of $\Aineqvecpluss$
to $\SI{10}{\%}$ of half the action space range
and its maximum value to half the action space range directly,
so that constraint satisfaction never becomes trivial
while guaranteeing a minimum exploration volume.

\subsection{Application}
\label{sec:application}

\begin{figure}[!t]
    \centering
    \includegraphics[width=0.40\columnwidth]{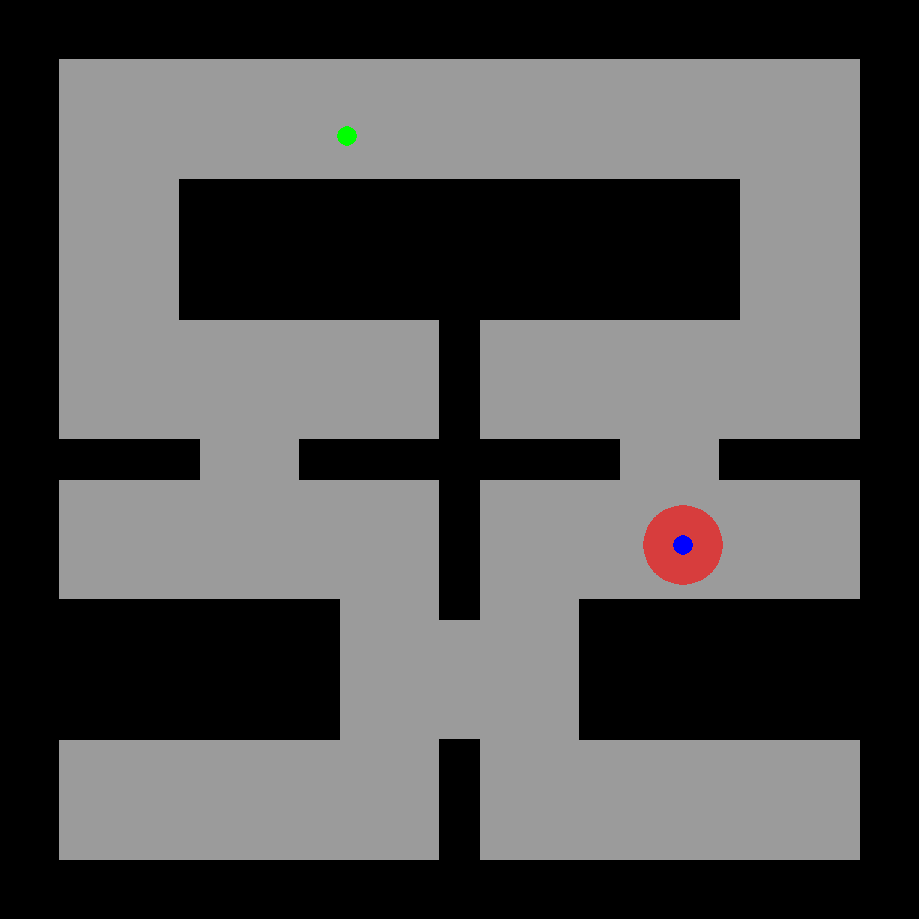}
    \hspace{0.5cm}
    \includegraphics[width=0.40\columnwidth]{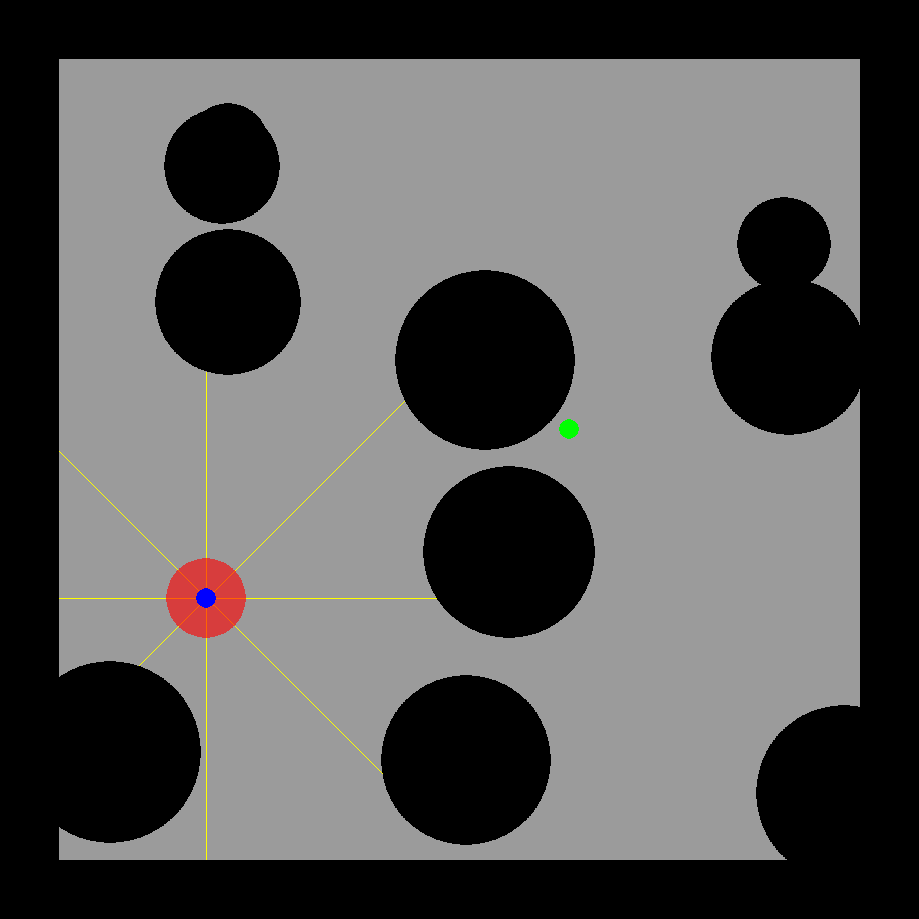}
    \caption{
        An agent (blue) is tasked to reach a target (green),
        while avoiding walls and obstacles (black)
        in a static maze (left) or with random obstacles (right)
        with shortest distances in eight directions (yellow).
        The agent can be controlled with position (within the red area)
        or force commands.
    }
    \label{fig:envs}
\end{figure}

We consider a task consisting in controlling an agent
to reach a target point in a maze-like environment,
see Fig.~\ref{fig:envs} (left).
Both the agent and the target are represented by circles of diameter $0.05$.
Throughout its motion, the agent has to avoid certain areas of the map:
the external bounds of the world
represented by a square of side 2,
$\Abrackets{-1, 1}^2$,
and holes in the ground, represented by black surfaces.
At the beginning of each episode,
agent and target positions are randomly sampled within the allowed surface.
At each timestep, the state vector comprises the position of the target and that of the agent.
The agent can then make a 2D motion as action vector
$\Aaction = \Aparentheses{\Delta x, \Delta y} \in \Abrackets{-0.1, 0.1}^{2}$.
If the norm of $\Aaction$ is within a maximum step size $\Delta_{m} = 0.1$,
the position is directly incremented by it,
else
it is clipped to lie within the circular movement range
of radius $\Delta_m$.
Each action thus results in an updated state and
a reward signal of the form
$\Areward = \Arewardfail + \Arewardsuccess + \Arewarddistance + \Arewardalive$,
with
$\Arewardfail = -10$ a penalty when reaching the border of the world or the central hole,
$\Arewardsuccess = +10$ a bonus when reaching the target,
$\Arewarddistance = -0.01*\Anorm{\mathbf{P}_T - \mathbf{P}}$ a reward on the distance between
the agent of the target (increasing towards zero as the distance decreases),
and
$\Arewardalive = -0.01$ a constant penalty per timestep encouraging rapid
completion of the task.
The episode ends when $\Ahorizon = 100$ timesteps have passed
or when either $\Arewardfail$
or $\Arewardsuccess$ occurs.

We then collect a set of expert demonstrations
by having a human user directly controlling the agent with the mouse,
without specific instructions on how to reach the goal (e.g., shortest path possible).
We collect $500$ such trajectories
and take them as our set of positive demonstrations $\Agooddemonstrations$.
As this environment does not involve complicated dynamics for the agent,
we can define bad actions as those immediately leading to fail the task.
We thus iterate through each positive demonstration
and sample actions along the circular action range of radius $\Delta_m$.
State-action couples leading to task failure are then taken as negative demonstrations.
As an additional heuristic, we also consider the expert path, reversed,
as negative demonstrations, i.e., if a positive demonstration
$\Aparentheses{\Aaction_i, \Astate_i}$ leads to the state $\Astate_{i+1}$,
then we take $\Aparentheses{\Astate_{i+1}, -\Aaction_i}$ as negative demonstration.
Note that these heuristics are only applicable due to the simplicity of the environment
and its dynamics.
We discuss their automatic discovery in the next Section.
Overall, we thus collect a set of $23222$ demonstrations:
$7228$ positive and $15994$ negative.

\begin{figure}[!t]
    \centering
    \subfloat[Loss and accuracy throughout constraint network training.] {
        \label{fig:cnet:training}
        \includegraphics[width=1.0\columnwidth]{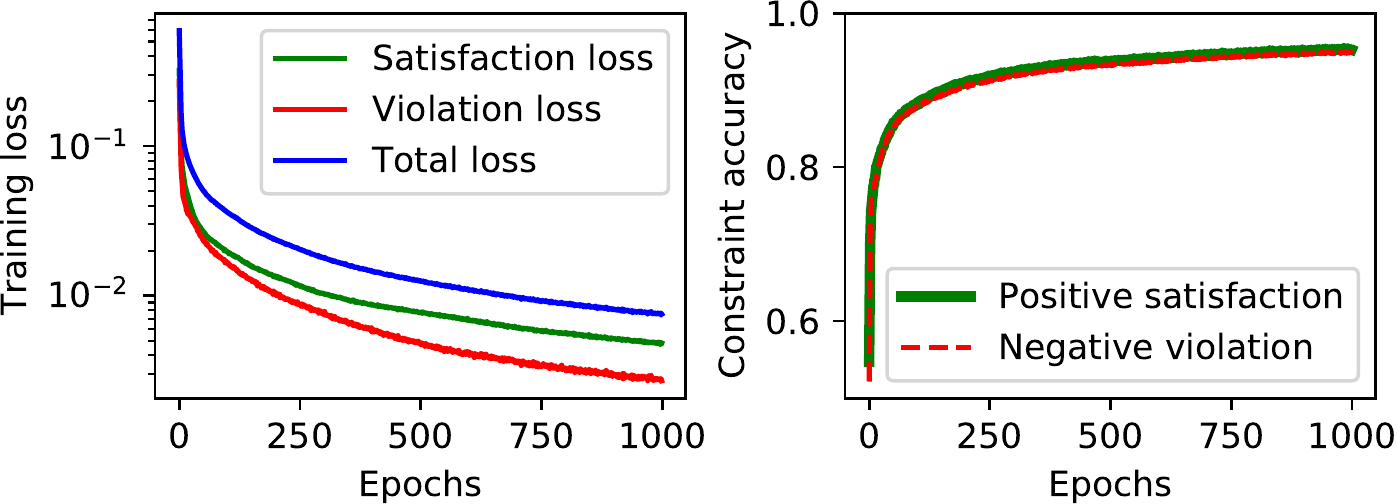}
    } \\
    \subfloat[Reinforcement learning with and without learned constraints.] {
        \label{fig:cnet:rl}
        \includegraphics[width=1.0\columnwidth]{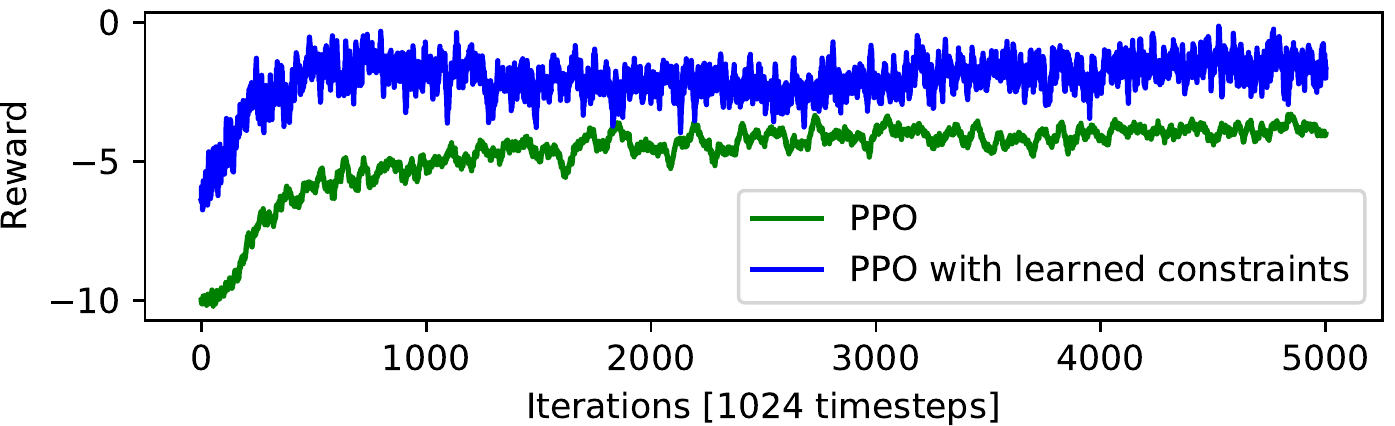}
    }
    \caption{
        Minimizing the constraint loss indeed results in correct separation
        for good and negative demonstrations (a).
        After training, the learned constraints can guide exploration
        during reinforcement learning to achieve higher rewards (b).
    }
    \label{fig:cnet}
\end{figure}

We then train a constraint network $\Acnet$ to predict
$\Anumineq = 2$ constraints on the action space of the agent when exploring the maze.
We minimize the constraint loss $\Acnetloss$
using the Adam optimizer, on mini-batches of size $64$ comprising
$32$ positive demonstrations
and $32$ negative demonstrations each.
In doing so, each training epoch consists in iterating through the
$15994$ negative demonstrations exactly once,
while each of the $7228$ positive demonstration appears on average $2.2$ times per epoch.
Alternatively, we could weigh violation and satisfaction losses differently in
Eq.~\eqref{eq:full_loss}, e.g., in function of their proportion in the total dataset.
We depict the resulting training loss in Fig.~\ref{fig:cnet:training}.
By counting how many positive (resp. negative)
demonstrations actually satisfy (resp. violate)
the predicted constraints
after each training epoch,
we empirically verify that the proposed loss $\Acnetloss$
constitutes a representative proxy to learn constraints from demonstrations only.
Once $\Acnet$ is done training,
we embed it within a reinforcement learning process
to predict constraints from states encountered during exploration
and thus guide the behavior of the agent.
Fig.~\ref{fig:cnet:rl} illustrates that this enables both starting
from higher rewards, since penalty-heavy collisions are avoided,
and reaching a higher reward after training.
We depict a full trajectory along with
a visualization of action-space constraints
in Fig.~\ref{fig:maze}.

\section{Constrained Exploration and Recovery from Experience Shaping}
\label{sec:ceres}

\subsection{Overview}
\label{sec:ceres:overview}

We established in
Section~\ref{sec:learning_constraints}
that
it is possible to learn action-space constraints as functions of states
to guide exploration during reinforcement learning,
given a set of positive and negative demonstrations.
However, the acquisition of such demonstrations is often problematic
on problems of practical interest.
First, one cannot always assume the availability of a human expert,
e.g., for tasks that humans struggle to complete and look to automate,
such as robotic tasks involving high payloads or requiring sub-millimeter accuracy.
Second, even when positive demonstrations are available,
there may not be clear heuristics to infer negative
from positive demonstrations
(e.g., by \enquote{reversing} them).
Third, direct sampling and success-failure evaluation
can quickly become intractable on high-dimensional state and action spaces.
Finally, even on low-dimensional domains,
one may not be able to evaluate an action in a single step.
Instead, the effects of a given action may only appear many steps later,
mitigated by other events that happened in between.
As a result,
it is essential to derive an algorithm
enabling the discovery and identification of positive and negative demonstrations
starting from scratch.

We propose to do so through the reinforcement learning setting.
First, we train a \emph{direct} control policy $\Adirectpolicynetwork$
that learns to complete the task.
After each trajectory,
state-action couples are evaluated to determine
if they can be labeled positive or negative.
In this framework,
we consider a demonstration as positive if from the successor step,
there exists a trajectory that does not lead to failure
within $\Anumrecoverysteps$ steps,
with $\Anumrecoverysteps$ a hyperparameter
to be chosen in function of dynamics of the task.
Conversely, we consider a demonstration as negative if
the resulting state only leads to failure within $\Anumrecoverysteps$.
At this stage, only the final demonstration can confidently be
labeled negative,
if the trajectory terminates with failure,
while only the first demonstrations,
of remaining trajectory length greater than $\Anumrecoverysteps$,
can confidently be labeled positive.

\subsection{Demonstration Sorting by Learning Recovery}
\label{sec:ceres:learning_recovery}

The second part of our algorithm thus consists
in transferring the uncertain demonstrations sampled by the direct policy
to a \emph{recovery} control policy $\Arecoverypolicynetwork$
that learns to recover from such uncertain states.
Namely, training of $\Arecoverypolicynetwork$
involves resetting episodes only to
uncertain states visited by the direct policy.
In addition, the reward signal $\Arewardrecovery$
used to train $\Arecoverypolicynetwork$ is simplified
to being equal to $+1$ if the agent is still alive at each timestep,
and $-\Anumrecoverysteps$ if it fails the task.
If the recovery agent is still active after $\Anumrecoverysteps$,
the demonstration leading to the episode's starting state (sampled from the direct policy)
is labeled as positive, recursively with all predecessor demonstrations.
Conversely, if recovery was unsuccessful for a chosen number of attempts
$\Anumrecoveryattempts$,
the starting direct demonstration is labeled as negative,
along with all successor demonstrations.
We remark that, when evaluating trajectories, it is useful to start from the middle
as the characterization of a given demonstration affects that of either all its
predecessors or all its successors, thus halving the search space each time.
Overall, the positive and negative demonstrations collected from both
direct and recovery policies can then be used to train a constraint network $\Acnet$
to guide the exploration for $\Adirectpolicynetwork$,
and optionally $\Arecoverypolicynetwork$.

In summary, given an environment $\Adirectenvironment$ on which
we seek to train a direct policy $\Adirectpolicynetwork$,
our approach necessitates the following adjustments 
in creating a recovery environment $\Arecoveryenvironment$ to train $\Arecoverypolicy$:
\begin{enumerate*}
    \item a simplified reward that only penalizes task failure,
    \item \label{item:ceres_requirement_flags} the availability of success and failure flags regarding the final action
        prior to episode termination, and
    \item \label{item:ceres_requirement_restore} a function restoring the environment to chosen states.
\end{enumerate*}
While \ref{item:ceres_requirement_restore} may appear rather restrictive,
the idea of restoring reference states was also used to guide reinforcement learning
for whole-body robot control
in~\cite{tog:peng:2018}.
Alternatively,
when such a restoration function is unavailable but
the environment can be finely controlled,
we could consider simply resetting it to reference states
by replaying a set number of demonstrations from the sampled direct trajectories.
Finally, \ref{item:ceres_requirement_flags} is necessary to confidently
classify the final demonstration, as early episode termination can occur
from reasons besides failure (negative),
such as completing the task (positive)
or just reaching a maximum number of timesteps (uncertain).

\subsection{Detailed Algorithm}

Conventionally, in the on-policy reinforcement learning setting,
states, actions, rewards and other relevant quantities
(e.g., value, termination, etc.)
are collected as
trajectories
$\Arltrajectory$
following predictions from the neural network policy
that are then executed onto the environment
in order to update a policy network $\Apolicynetwork$,
through the use of a $\textsc{UpdatePolicy}$ method
e.g., PPO.
In CERES, described in Fig.~\ref{alg:ceres},
positive, negative, and uncertain demonstrations are sampled together with
$\Arltrajectory$ within a $\textsc{Sample}$ method.
Each time a state-action demonstration $\Apsa$
is labeled as positive or negative,
we store it together with the associated indicator $\Aindicatorsa$,
into an experience replay buffer $\Aexperiencereplaybuffer$,
then used to
iteratively train a constraint network $\Acnet$
with an $\textsc{UpdateConstraints}$ method
following Section~\ref{sec:learning_constraints}.
In parallel with each policy update,
Uncertain trajectories are also transfered from direct to recovery environments
to serve as episode initialization states.

\begin{figure}[!t]
    \begin{algorithmic}[1]
        \Procedure{CERES}{$\Adirectenvironment, \Arecoveryenvironment, \Adirectpolicynetwork, \Arecoverypolicynetwork, \Acnet$}
        \State $\Adirectpolicynetwork$.initialize(), $\Arecoverypolicynetwork$.initialize(), $\Acnet$.initialize()
        \State $\Aexperiencereplaybuffer = \Abraces{}$ \Comment Experience replay buffer start empty
        \For{$i_\text{iter} = 1$ to $n_\text{iter}$}
        \State $\Arltrajectory^{d}, D_+^d, D_-^d, D_*^d =
            \textsc{Sample}{\Aparentheses{\Adirectenvironment, \Adirectpolicynetwork, \Acnet}}$
        \State $\textsc{UpdatePolicy}{\Aparentheses{\Adirectpolicynetwork, \Arltrajectory^{d}}}$
        \State $\tau_\text{RL}^{r}, D_+^r, D_-^r, D_*^r =
            \textsc{Sample}{\Aparentheses{\Arecoveryenvironment, \Arecoverypolicynetwork, \Acnet}}$
        \State $\textsc{UpdatePolicy}{\Aparentheses{\Arecoverypolicynetwork, \Arltrajectory^{r}}}$
        \State $\Aexperiencereplaybuffer$.append($D_+^d, D_-^d, D_+^r, D_-^r$)
        \State $\textsc{UpdateConstraints}{\Aparentheses{\Acnet, \Aexperiencereplaybuffer}}$
        \State $\Arecoveryenvironment$.add\_for\_recovery${\Aparentheses{D_*^d}}$
        \EndFor
        \State \textbf{return} trained $\Adirectpolicynetwork, \Arecoverypolicynetwork, \Acnet$ 
        \EndProcedure
    \end{algorithmic}
    \caption{
        In CERES, a direct control policy is trained together with a recovery policy,
        yielding positive and negative demonstrations to learn
        and apply action-space constraints.
    }\label{alg:ceres}
\end{figure}

The $\textsc{Sample}$ method is further described in
Fig.~\ref{alg:sample}.
We highlight in particular the following.
On line~\ref{alg:sample:correct},
raw actions predicted by the policy network are corrected using
a method $\textsc{Constrain}$ implementing the quadratic program of Eq.~\eqref{eq:qp}.
Then, on line~\ref{alg:sample:rltrajectory},
it is the initial prediction that is used for policy update
and not the corrected action, as training is done on-policy.
Still, on line~\ref{alg:sample:cnettrajectory}
it is the corrected action that is used as reference demonstration,
since it is the action that is effectively performed onto the environment.
Finally,
given such unlabeled demonstrations,
we sort them as positive, negative and uncertain through a procedure
$\textsc{EvaluateDemos}$ implementing the logic described
in Section~\ref{sec:ceres:learning_recovery}.

\begin{figure}[!t]
    \begin{algorithmic}[1]
        \Procedure{Sample}{$\Aenvironment, \Apolicynetwork, \Acnet$}
        \State $\Arltrajectory = ()$ \Comment{Trajectories for policy update}
        \State $\Acnettrajectory  = ()$ \Comment{Trajectories for demonstration evaluation}
        \State $\Astate = \Aenvironment$.reset(); $\text{end} = \text{false}$ \Comment{Get initial state}
        \While{not end}
        \State $S$.add($\Astate$) \Comment{Store current state}
        \State $\Aaction = \Apolicynetwork(\Astate)$ \Comment{Predict action}
        \State $\Aineqmats, \Aineqvecs = \Acnet(\Astate)$ \Comment{Predict constraints}
        \State $\Acorrectedaction = \textsc{Constrain}{\Aparentheses{\Aaction, \Aineqmats, \Aineqvecs}}$ \Comment{Correct action}
        \label{alg:sample:correct}
        \State $\widetilde{\Astate}, \Areward, \text{end}, \text{info} = \Aenvironment.$do${\Aparentheses{\Acorrectedaction}}$ \Comment{Play corrected}
        \State $\Arltrajectory$.append${\Aparentheses{\Astate, \Aaction, \Areward, \text{end}}}$
        \label{alg:sample:rltrajectory}
        \State $\Acnettrajectory$.append${\Aparentheses{\Astate, \Acorrectedaction, \text{info}}}$
        \label{alg:sample:cnettrajectory}
        \EndWhile
        \State $D_+, D_-, D_* = \textsc{EvaluateDemos}{\Aparentheses{\Acnettrajectory}}$ 
        \State \textbf{return} $\Arltrajectory, D_+, D_-, D_*$ 
        \EndProcedure
    \end{algorithmic}
    \caption{
        Trajectories are sampled for policy and constraint learning
        from
        positive, negative, uncertain demonstrations.
    }\label{alg:sample}
\end{figure}

\section{Experiments}
\label{sec:experiments}

\subsection{Practical Implementation}

We implement the constraint learning framework and the CERES algorithm
within Tensorflow,
while building upon the OpenAI Baselines with PPO
as reinforcement learning method for training direct and recovery agents.
Preliminary experiments showed that
since constraint predictions can be rather inaccurate over the first iterations,
as labeled demonstrations are still few,
it is possible to only correct the action prediction 
with a certain probability in Fig.~\ref{alg:sample}, line~\ref{alg:sample:correct},
and otherwise play the predicted action directly in the environment.
We empirically found that an appropriate metric
for the constraint activation probability is
the percentage of actions that are
correctly separated by the predicted constraints
(i.e., the proportion of positive actions satisfying the predicted constraints
and negative actions violating them).
We also obtained good results by only constraining the direct policy,
enabling a more diverse range of sampled actions to learn recovery,
prior to training the constraint network.

\subsection{Obstacle Avoidance with Dynamics}

While the example considered in Section~\ref{sec:application}
was limited by fixed safe domains and position control for the agent,
we now consider the case where hole placement is randomized at each episode,
see Fig.~\ref{fig:envs} (right),
and where the agent can be controlled with force commands.
In the latter case, it is now insufficient to evaluate demonstrations as good or bad
from a single step,
as the agent can no longer stop instantly if it is travelling at
maximum speed.
In addition to the positions of the agent and the target,
the state vector now includes the current linear velocity of the agent
and its distance to surrounding obstacles
akin to LIDAR systems,
along
$8$ regularly-spaced beams starting from its center.

\begin{figure}[!t]
    \centering
    \includegraphics[width=1.0\columnwidth]{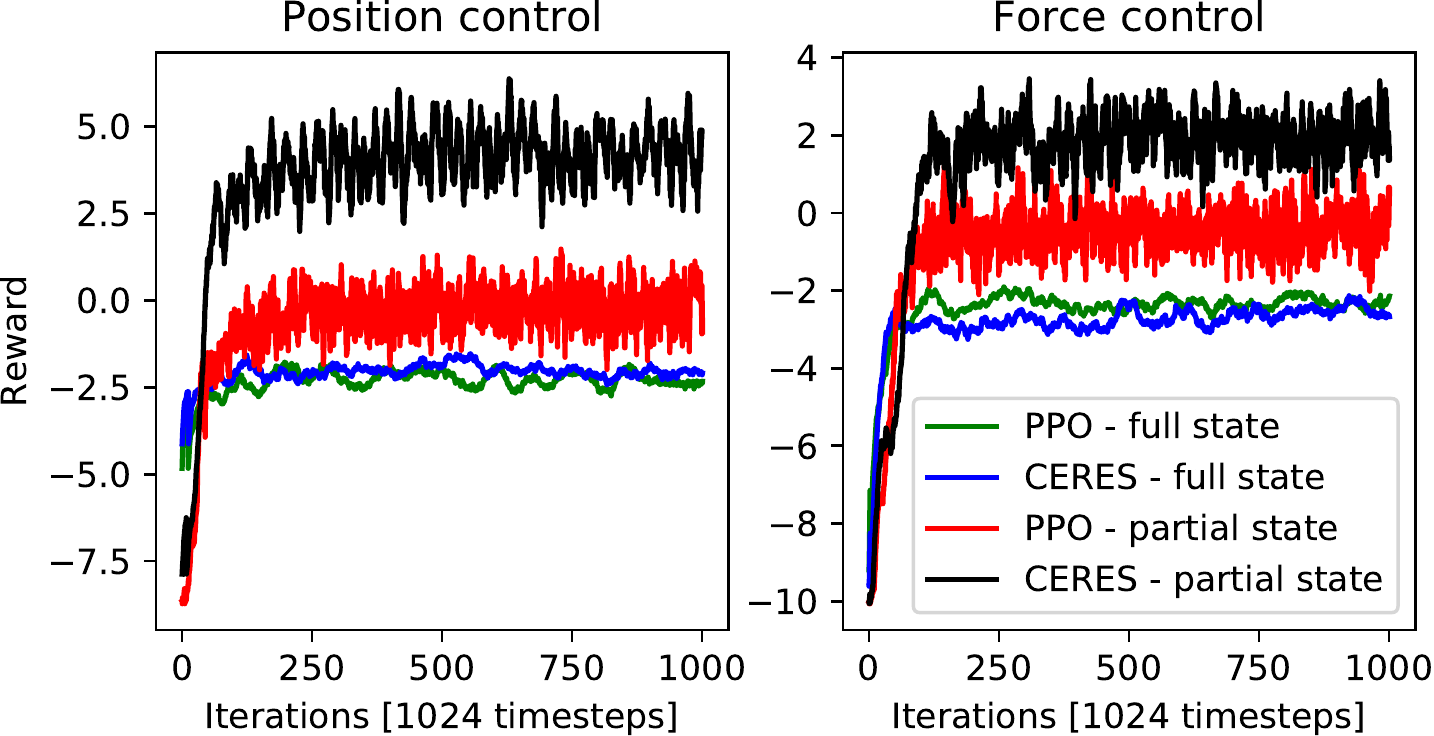}
    \caption{
        Reinforcement learning on random obstacles.
    }
    \label{fig:obstacles}
\end{figure}

We consider four variants of this environment:
two where the agent is controlled in position, with 2D position increments as actions,
and two where it is controlled with 2D forces as actions,
updating its velocity and position by consecutive integration.
For each control setting,
we consider the case where all observations are provided to the control policy,
i.e., agent, target and obstacle informations,
and the case where the control policy has only access
to agent and target information,
while the constraint network still has access to the whole state vector.
We evaluate CERES against vanilla PPO,
sharing the same reinforcement learning hyperparameters and random seeds
and depict the resulting rewards in Fig.~\ref{fig:obstacles}.
Overall, while full-state tasks seem difficult to achieve in the first place,
CERES enables the safe learning from fewer observations.
Indeed, in such situations, considerations of distances can be left to the 
constraint network while the policy network can focus on general navigation.

\section{Discussion and future work}
\label{sec:conclusion}

In our work, we established that
expert demonstrations could be used in a novel way,
to learn safety constraints from positive and negative examples.
When both are available, the resulting constraints
can accelerate reinforcement learning by starting from
and reaching higher rewards.
Towards applications of practical interest,
we derived a new algorithm, CERES,
enabling the automatic discovery of such positive and negative examples,
and thus the learning of safety constraints from scratch.
On a task involving multi-step dynamics, we demonstrated that
our approach could preserve such advantages in terms of rewards,
while also enabling the main control policy to learn from fewer observations.
Possible future developments include
tackling real-world robotics applications
and problems where success and failure metrics
are more ambiguous.

%%%%%%%%%%%%%%%%
% Bibliography %
%%%%%%%%%%%%%%%%

\bibliographystyle{aaai}
\bibliography{extra/actshape}

\end{document}